\documentclass[sigconf]{acmart}
\renewcommand\footnotetextcopyrightpermission[1]{}

\usepackage{subcaption}
\usepackage{multirow}
\usepackage{algorithm}
\usepackage{algpseudocode}
\usepackage{enumitem}
\AtBeginDocument{%
  }

\acmConference[ICMR '26]{ACM International Conference on Multimedia Retrieval}
{June 16-19, 2026}{Amsterdam, The Netherlands}

\acmSubmissionID{0803}

\begin{document}

% \title{UML-CoT: Structured Reasoning and Planning\\ with Unified Modeling Language for Robotic Room Cleaning}
\title{OOWM: Structuring Embodied Reasoning and Planning via Object-Oriented Programmatic World Modeling}


\author{
  Hongyu Chen\textsuperscript{1},
  Liang Lin\textsuperscript{1,2,3},
  Guangrun Wang\textsuperscript{1,2,3,*}
  \\
  \small{
    \textbf{Email:} chehy527@mail2.sysu.edu.cn, linliang@live.com, wanggrun@gmail.com
  }
}

\affiliation{
    \institution{
        \textsuperscript{1}Sun Yat-sen University; \textsuperscript{2}Guangdong Key Laboratory of Big Data Analysis and Processing;\textsuperscript{3}X-Era AI Lab
    }
    \city{}
    \country{}
}

\renewcommand{\shortauthors}{Chen et al.}


\begin{abstract}
Standard Chain-of-Thought (CoT) prompting empowers Large Language Models (LLMs) with reasoning capabilities, yet its reliance on linear natural language is inherently insufficient for effective world modeling in embodied tasks. While text offers flexibility, it fails to explicitly represent the state-space, object hierarchies, and causal dependencies required for robust robotic planning. To address these limitations, we propose Object-Oriented World Modeling (OOWM), a novel framework that structures embodied reasoning through the lens of software engineering formalisms. 
We redefine the world model not as a latent vector space, but as an explicit symbolic tuple $\mathcal{W} = \langle \mathcal{S}, \mathcal{T} \rangle$: a \textbf{State Abstraction} ($G_\text{state}$) instantiating the environmental state $\mathcal{S}$, coupled with a \textbf{Control Policy} ($G_\text{control}$) representing the transition logic $\mathcal{T}: S \times A \rightarrow S'$.
OOWM leverages the Unified Modeling Language (UML) to materialize this definition: it employs Class Diagrams to ground visual perception into rigorous object hierarchies, and Activity Diagrams to operationalize planning into executable control flows.
Furthermore, we introduce a three-stage training pipeline combining Supervised Fine-Tuning (SFT) with Group Relative Policy Optimization (GRPO). Crucially, this method utilizes outcome-based rewards from the final plan to implicitly optimize the underlying object-oriented reasoning structure, enabling effective learning even with sparse annotations. Extensive evaluations on the MRoom-30k benchmark demonstrate that OOWM significantly outperforms unstructured textual baselines in planning coherence, execution success, and structural fidelity, establishing a new paradigm for structured embodied reasoning.
\end{abstract}



\begin{teaserfigure}
  \includegraphics[width=\textwidth]{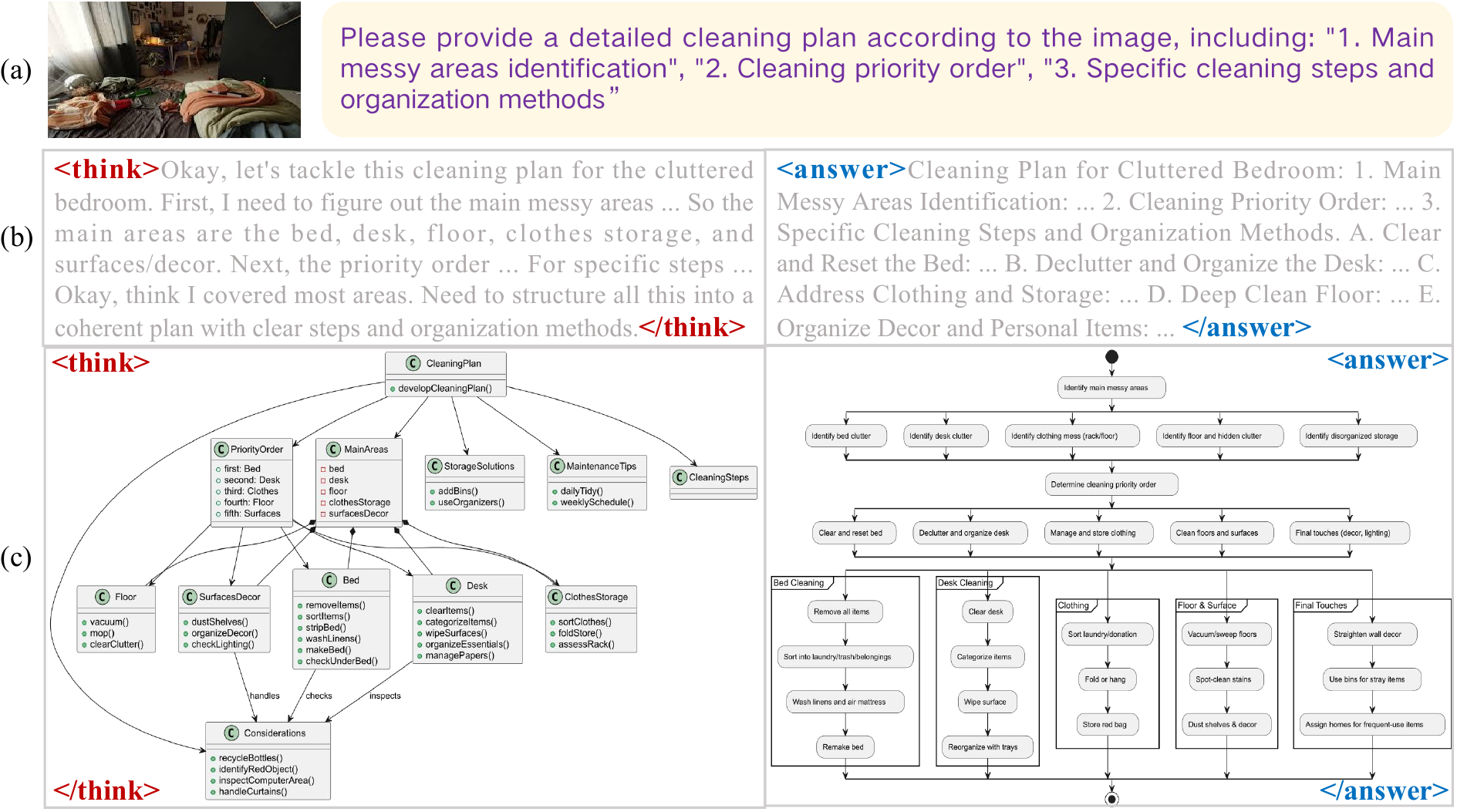}
  \caption{\textbf{Comparison of Standard Text-based CoT vs.\ Object-Oriented Programmatic World Modeling (OOWM).}
    (a) \textbf{Input:} A visual observation of a cluttered room together with a specific cleaning instruction.
    (b) \textbf{Unstructured Textual CoT:} The baseline model produces linear natural-language reasoning, resulting in a shallow world model that lacks explicit state definitions and executable control logic.
    (c) \textbf{Our OOWM Framework:} The agent structures its reasoning by first constructing a \textbf{State Abstraction} (UML Class Diagram) that defines object hierarchies and properties, followed by a \textbf{Control Policy} (UML Activity Diagram) that governs sequential execution.
    This shift from free-form text to standardized object-oriented modeling enforces rigorous plan coherence and enables direct executability.
    \emph{Please zoom in to view details clearly.}}
  \label{fig:Intro}
\end{teaserfigure}


\maketitle

\begingroup
  \renewcommand\thefootnote{\fnsymbol{footnote}} % use *, †, ‡, …
  \footnotetext[2]{Corresponding author: Guangrun Wang}
\endgroup

\section{Introduction}\label{sec:intro}
Embodied AI systems, particularly those operating in cluttered, real-world environments, face a fundamental challenge in representation: they must bridge the gap between high-level reasoning and low-level physical actuation \citep{zitkovich2023rt,brohan2022rt,black2024pi_0,intelligence2504pi0,zhan2025mathcal,zhan2026stable,li2025vla,song2025physical,chen2026radar}. Successful execution requires a robust internal \textbf{world model}—a structured understanding of object properties, spatial hierarchies, and causal action dependencies. While recent advancements in Large Language Models (LLMs) have enabled agents to generate reasoning traces through \textit{Chain-of-Thought} (CoT) prompting \citep{DBLP:conf/nips/Wei0SBIXCLZ22, DBLP:conf/nips/KojimaGRMI22}, reliance on unstructured natural language remains a critical bottleneck for embodied planning \citep{li2025situ,xu2026bridging}.

The primary limitation of \textbf{text-based CoT} is its inherent linearity, which conflicts with the multi-dimensional nature of physical environments. Textual reasoning often results in ``shallow world models'' \citep{DBLP:journals/tmlr/0001Z00KS24, DBLP:conf/acl/WangXLHLLL23} that lack explicit structures for modeling object states or action preconditions. This representation gap leads to three specific failures: (i) ambiguity in distinguishing between an object's static attributes and its dynamic states; (ii) difficulty in verifying logical consistency across long-horizon plans \citep{DBLP:conf/iclr/CreswellSH23}; and (iii) a lack of executable formalisms, requiring additional translation steps to convert reasoning into robotic control policies. To address these issues, prior works have explored intermediate symbolic representations, such as scene graphs or logic formulations \citep{DBLP:conf/emnlp/PanAWW23,DBLP:conf/aaai/BestaBKGPGGLNNH24,DBLP:conf/nips/YaoYZS00N23}.

However, \textbf{traditional graph-based reasoning} remains insufficient for comprehensive world modeling. While graphs capture binary or ternary relations, they lack the expressive power of \textit{object-oriented} design—specifically, the ability to model inheritance, aggregation, and behavioral abstraction. Furthermore, standard graph representations lack standardized semantics for procedural control, making it difficult to encode the sequential, conditional, and iterative logic required for robust cleaning plans. Consequently, existing methods often rely on ad hoc, task-specific graph definitions that do not generalize across domains.

To bridge this gap, we propose \textbf{Object-Oriented World Modeling (OOWM)}, a paradigm that treats the environment not as a sequence of words or a web of nodes, but as a system of interacting objects and processes. 
In this framework, we formally define the world model as a dual-component symbolic architecture: a State Abstraction ($G_\text{state}$) that maps observations to a structured state space $S$, and a \textit{Control Policy} ($G_{control}$) that approximates the transition function $T: S \times A \rightarrow S'$ governing future states. 
We operationalize this paradigm using the Unified Modeling Language (UML)—a standardized formalism from software engineering \citep{DBLP:journals/jot/Ashbacher04a}. UML uniquely satisfies the requirements of embodied planning: it provides \textbf{Class Diagrams} to construct a State Abstraction (capturing object hierarchies and attributes) and \textbf{Activity Diagrams} to define control policies (encoding executable plans with flow control). By adopting this formalism, we transform the reasoning process from a ``stream of consciousness'' into a structured architectural design.

Building on this philosophy, we introduce the \textbf{OOWM Framework} for structured embodied reasoning. The agent first perceives the environment and constructs a \textbf{UML Class Diagram}, effectively instantiating a symbolic object-oriented world model. It then derives a cleaning strategy in the form of a \textbf{UML Activity Diagram}, ensuring that the plan is both logically sound and directly executable.
To train this capability, we introduce a \textbf{three-stage learning strategy}: (1) Supervised Fine-Tuning (SFT) to initialize the model's ability to generate valid UML syntax and reasoning structures; (2) Reinforcement Learning Fine-Tuning (RLFT) using Group Relative Policy Optimization (GRPO) \citep{DBLP:journals/corr/abs-2402-03300}, where the model is rewarded based on the semantic correctness of the final plan; and (3) Answer-Only GRPO, which optimizes the intermediate reasoning structure implicitly via outcome-based rewards.

We evaluate our framework on the \textbf{MRoom-30k} dataset, a new benchmark simulating diverse, cluttered room scenarios. Empirical results show that by structuring reasoning through Object-Oriented World Modeling, our approach significantly outperforms unstructured textual baselines in plan coherence, executability, and structural fidelity. Fig.~\ref{fig:Intro} illustrates the contrast between unstructured CoT and our OOWM approach in robotic room cleaning.

\textbf{Our contributions are:}
(1) The proposal of \textbf{Object-Oriented World Modeling (OOWM)}, a framework that unifies symbolic state representation with executable planning using standardized UML formalisms;
(2) A three-stage training pipeline leveraging GRPO to optimize structured reasoning through outcome-based reinforcement;
(3) The introduction of \textbf{MRoom-30k}, a large-scale benchmark of cluttered indoor environments annotated for structured reasoning tasks;
(4) Empirical evidence demonstrating that standardized software engineering formalisms provide superior interpretability and reliability compared to text-based and graph-based baselines.

\section{Related Work}

\noindent\textbf{Chain-of-Thought Reasoning and its Limits.}\quad
Chain-of-Thought (CoT) prompting has revolutionized LLM reasoning by decomposing complex problems into intermediate intermediate steps \citep{DBLP:conf/nips/Wei0SBIXCLZ22}. Numerous variants have emerged to enhance this process, including Self-Consistency \citep{DBLP:conf/iclr/0002WSLCNCZ23} for robustness, Least-to-Most Prompting \citep{DBLP:conf/iclr/ZhouSHWS0SCBLC23} for problem decomposition, and iterative refinement strategies like STaR \citep{DBLP:conf/nips/ZelikmanWMG22}.
However, a fundamental limitation persists across these methods: they treat reasoning as a linear, unstructured stream of natural language tokens. While recent efforts like Semi-Structured CoT \citep{DBLP:conf/naacl/SuLBH24} and Faithful Logical CoT \citep{DBLP:conf/acl/Xu0P0LH24} introduce auxiliary structural signals, they typically rely on shallow graph representations that lack semantic depth.
Our work departs from this linear paradigm. We propose Object-Oriented World Modeling (OOWM), which redefines reasoning not as text generation, but as the \textit{instantiation of a structured system}. By implementing this paradigm via UML, we shift the reasoning process from transient narration to rigorous architectural modeling, capturing both state ($G_\text{state}$) and behavior ($G_\text{control}$) in a unified framework.

\noindent\textbf{Structured World Modeling in Embodied AI.}\quad
To ground LLMs in physical environments \citep{chen2025style4d,xiang2025distilled,zou2024seconds}, recent works have adopted graph-based symbolic representations. Scene graphs \citep{DBLP:journals/corr/abs-2408-16098} and logic graphs \citep{DBLP:conf/acl/Xu0P0LH24} are commonly used to map static object relations, while neuro-symbolic planners like SymPlanner \citep{DBLP:journals/corr/abs-2505-01479} and PDDL-based translators \citep{DBLP:journals/corr/abs-2503-17309, DBLP:conf/rss/HanZZWZ24} attempt to formalize action sequences.
However, traditional graph-based approaches suffer from limited expressivity: they primarily model binary node-edge relations and struggle to capture higher-order concepts such as inheritance, encapsulation, and complex procedural flow (e.g., loops and conditional branching).
To overcome these representational bottlenecks, we operationalize OOWM using the Unified Modeling Language (UML). Unlike ad-hoc graphs, UML provides a standardized ontology for world modeling: Class Diagrams allow for a precise definition of the \textit{State Abstraction} (including object attributes and hierarchies), while Activity Diagrams formally encode the \textit{Control Policy}. This distinction allows our framework to move beyond simple ``relation extraction'' toward a comprehensive simulation of agent-environment interactions.

\noindent\textbf{Reinforcement Learning for Structured Reasoning.}\quad
Aligning LLM reasoning with task objectives often requires optimization beyond standard supervised learning. Techniques such as InstructGPT \citep{DBLP:conf/nips/Ouyang0JAWMZASR22} and RRHF \citep{DBLP:journals/corr/abs-2304-05302} utilize reinforcement learning (RL) to align model outputs with human preferences. More recently, Group Relative Policy Optimization (GRPO) \citep{DBLP:journals/corr/abs-2402-03300} has demonstrated that reasoning capabilities can be improved by propagating rewards from final outcomes to intermediate steps, even without dense step-by-step annotations.
We adapt these insights into a three-stage OOWM training pipeline. Since constructing valid UML world models requires rigorous syntax and semantic logic, we first utilize Supervised Fine-Tuning (SFT) for initialization. We then employ outcome-based GRPO to implicitly optimize the underlying world model structure. This ensures that the agent learns to construct high-fidelity Class and Activity diagrams not just by mimicking syntax, but by maximizing the execution success of the derived plans, effectively bridging the gap between symbolic structure and physical actuation.

\section{Methodology}

\subsection{Task Definition: Object-Oriented World Modeling}
\label{task}
We formulate the embodied planning challenge not merely as a sequence-to-sequence text generation task, but as a \textbf{System Modeling} problem. While conventional Chain-of-Thought (CoT) approaches attempt to bridge perception and action via linear, unstructured natural language, our framework—Object-Oriented World Modeling (OOWM)—structures this process by explicitly separating the representation of the environment's state from the agent's control logic.

We formally define the embodied World Model $\mathcal{W}$ as a symbolic tuple $\mathcal{W} = \langle \mathcal{S}, \mathcal{T} \rangle$. Here, $\mathcal{S}$ denotes the \textbf{State Abstraction}, which maps high-dimensional sensory inputs (images) into a structured object system, explicitly defining entity hierarchies, attributes, and static relationships. Complementing this, $\mathcal{T}$ denotes the \textbf{Transition Logic}, encoding the causal rules and control flows—including sequences, branches, and loops—that govern how agent actions transform the environment state.

Given a visual observation $x$ of a cluttered environment, the objective is to learn a mapping $f: x \rightarrow \mathcal{W}$. The OOWM paradigm mandates that the agent instantiates this tuple through two coupled symbolic components: 
First, it constructs the \textbf{State Abstraction} ($G_\text{state}$), which serves as the concrete instantiation of the State $\mathcal{S}$. This component functions as the reasoning foundation, utilizing the syntax of \textbf{UML Class Diagrams} within \texttt{<think>} tags to represent the scene's static semantics. 
Subsequently, the agent derives \textbf{The Control Policy} ($G_{control}$), which serves as the concrete instantiation of the Transition Logic ($\mathcal{T}$). This component acts as the executable plan, employing the syntax of \textbf{UML Activity Diagrams} within \texttt{<answer>} tags to operationalize the cleaning strategy into a verifiable workflow.

\begin{figure}
    \centering
    \includegraphics[width=1\linewidth]{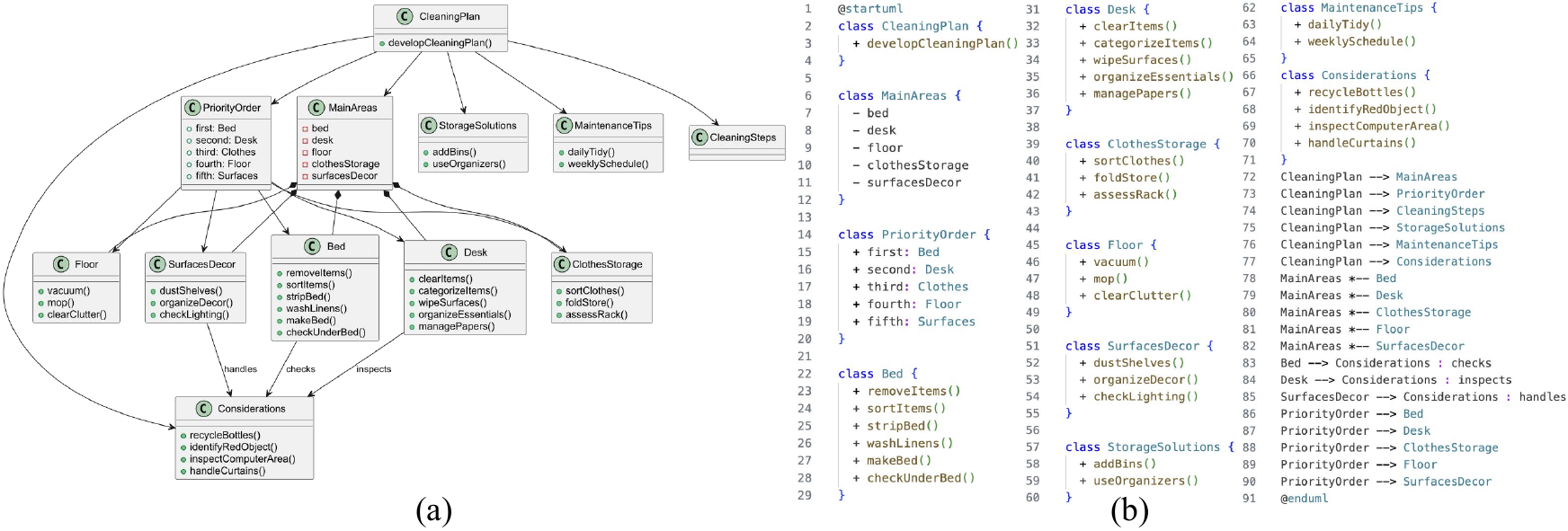}
    \vspace{-18pt}
    \caption{Instantiation of the \textbf{State Abstraction ($G_\text{state}$)}. (a) The visual abstraction rendered as a UML Class Diagram, defining object hierarchies and attributes. (b) The corresponding \textbf{serialized PlantUML code}, which serves as the model's actual symbolic output for reasoning. \emph{Please zoom in to view details clearly.}}
    \label{fig:think_con}
\end{figure}

\begin{figure}
    \centering
    \includegraphics[width=1\linewidth]{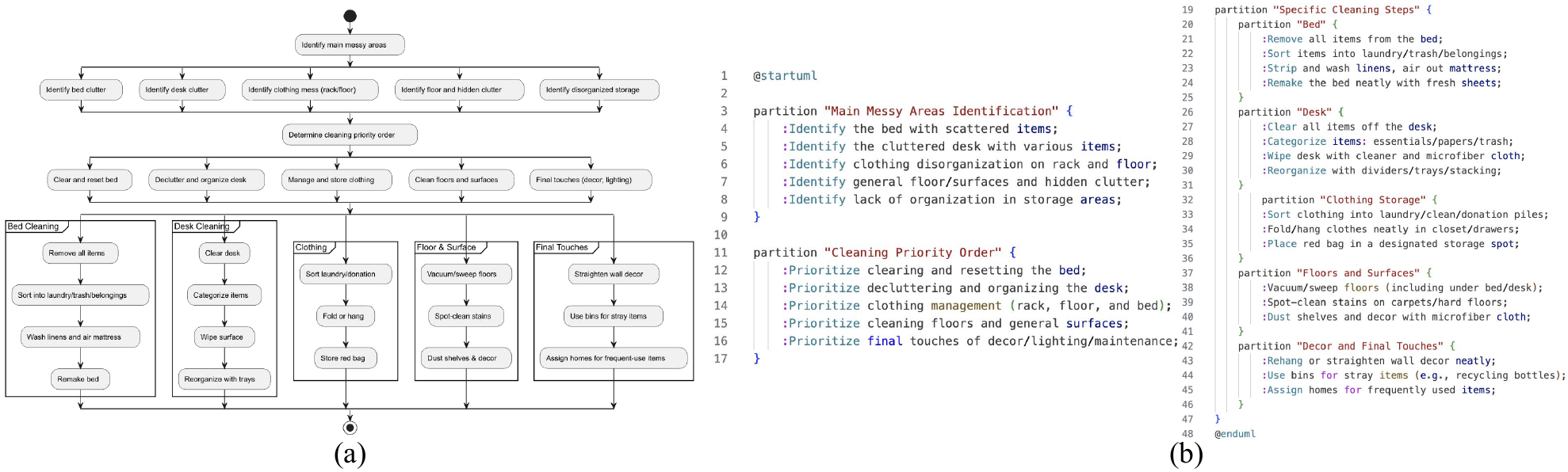}
    \vspace{-18pt}
    \caption{Instantiation of the \textbf{Control Policy ($G_{control}$)}. (a) The visual abstraction rendered as a UML Activity Diagram, illustrating the executable control flow. (b) The corresponding \textbf{serialized PlantUML code}. \emph{Please zoom in to view details clearly.}}
    \label{fig:answer_con}
\end{figure}

It is important to note that the agent's actual output is the PlantUML source code. This textual code serves as a serialized definition of OOWM components, which can be deterministically rendered into visual diagrams using external tools. Fig.~\ref{fig:think_con} and Fig.~\ref{fig:answer_con} explicitly demonstrate this equivalence, showing that a UML diagram and its PlantUML source are two interchangeable representations of the same structure.

\begin{figure*}[t]
    \centering
    \includegraphics[width=1\linewidth]{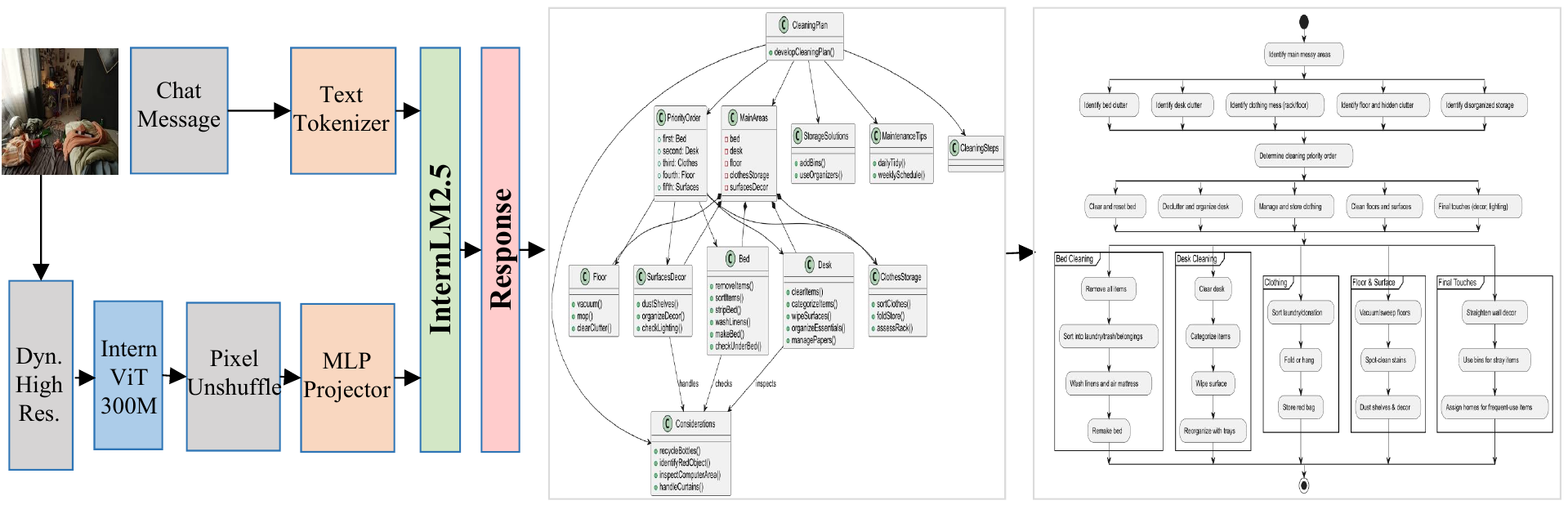}
    \caption{Overview of the proposed OOWM architecture. The input image is processed via dynamic resolution slicing and encoded by InternViT-300M, fused with textual prompts. The language decoder (InternLM2.5) functions as an \textbf{OOWM Instantiator}. Instead of free-form text, it synthesizes \textbf{serialized symbolic code} to construct two coupled modeling components: a \textbf{State Abstraction ($G_\text{state}$)} for grounding environmental semantics (reasoning), and a \textbf{Control Policy ($G_\text{control}$)} for governing the executable cleaning workflow (planning). \emph{Please zoom in to view details clearly.}}
    \label{fig:architecture}
\end{figure*}

To evaluate the quality of generated plan, we define a semantic similarity metric between the predicted control policy ($G_{\text{control}}$) and its ground-truth reference. Crucially, this metric serves as the primary reward signal during the reinforcement learning stage. Further details are provided in Section \ref{training_strategy}.

\subsection{Dataset Construction}
\label{Dataset_construction}

Existing indoor scene datasets, such as the MIT Indoor Scenes dataset \citep{DBLP:conf/cvpr/QuattoniT09}, suffer from a pronounced cleanliness bias—featuring predominantly tidy environments and lacking sufficient coverage of cluttered or disorganized household settings. More critically, they lack the structural logic annotations required to train rigorous world modeling abilities. To address these limitations, we introduce the \textbf{MRoom-30k}, a large-scale benchmark designed to evaluate Object-Oriented World Modeling in cluttered real-world scenarios.

The dataset consists of 30,792 images sourced from diverse platforms (Google, Bing, Baidu, Rednote) and the Messy Rooms Dataset \citep{DBLP:conf/nips/BhalgatLHVZ23}. These images span various household environments and exhibit varying levels of messiness, providing a rich visual basis for embodied planning tasks.

\noindent \textbf{Hierarchical OOWM Annotation.}\quad To support the training of the mapping $f:x\rightarrow \mathcal{W}$, we constructed the dataset with a hierarchical supervision structure using GPT-4o as the expert oracle. The annotations are stored in serialized PlantUML format and divided into two distinct subsets. 
First, for \textbf{Reasoning-Enhanced Subset} (1,000 samples), the expert explicitly constructs the \textbf{State Abstraction ($G_\text{state}$)}—instantiating the state $\mathcal{S}$ as a UML Class Diagram—before deriving the final \textbf{Control Policy ($G_\text{control}$)}. This ensures the agent learns to ground visual inputs into object hierarchies before planning actions.
Second, for the larger \textbf{Base Planning Set} (~29k samples), we focus on scaling up learning via outcome-based reinforcement learning (Stage 3). Here, the annotation is concentrated solely on the \textbf{Transition Logic ($\mathcal{T}$)}, providing the ground-truth \textbf{Control Policy ($G_\text{control}$)}. The expert generates detailed cleaning plans covering messy area identification, cleaning priority, and step-by-step actions, formalized as partitions within a UML Activity Diagram.

\noindent \textbf{Unstructured Baseline Annotation.}\quad To facilitate a rigorous comparison, we also provide parallel unstructured textual annotations for both subsets. These contain the same semantic content but are expressed in linear natural language, serving as the ground truth for traditional Text-CoT baselines.

Consequently, MRoom-30k offers a dual-format corpus: (i) Text Representation for benchmarking standard LLM approaches, and (ii) UML Representation (serialized as PlantUML code) for validating the effectiveness of our object-oriented world modeling paradigm. 

\subsection{Model Architecture and I/O Representation}
We adopt \textbf{InternVL 2.5} \citep{chen2024expanding, wang2024mpo} as the backbone for our structured multimodal reasoning framework. InternVL is a state-of-the-art vision-language model that integrates a visual encoder and a language decoder in a unified architecture, enabling effective grounding between image content and symbolic reasoning.

Each input instance consists of a single image depicting a cluttered room. To preserve both global context and local detail-crucial for identifying small objects in mess-InternVL applies a dynamic resolution slicing strategy. This divides the image into fixed-size patches while retaining a resized global view. The visual features are processed through a pixel unshuffle and MLP projector before being fused with the tokenized text prompt to form a joint multimodal input.

The core innovation lies in the decoder's output representation. The language decoder (InternLM 2.5) is optimized to function as an OOWM Instantiator. Instead of generating unstructured natural language, it synthesizes serialized symbolic code (in PlantUML syntax) to explicitly construct the World Model tuple $\mathcal{W} = \langle \mathcal{S}, \mathcal{T} \rangle$. 
For each image, the model sequentially produces two coupled components. First, it generates the State Abstraction ($G_\text{state}$), mapping visual features to a structured object hierarchy. Subsequently, it derives the Control Policy ($G_\text{control}$), which instantiates the Transition Logic ($\mathcal{T}$), governing the executable cleaning workflow.

This architecture (Fig. \ref{fig:architecture}) enables the joint modeling of visual perception and object-oriented reasoning, producing interpretable outputs that bridge the gap between scene understanding and structured action generation.

\subsection{Multi-Stage Training Strategy}
\label{training_strategy}

To equip the agent with the capability to perform Object-Oriented World Modeling(OOWM) in complex environments, we propose a three-stage training strategy. This pipeline progressively enhances the model's performance, evolving from mimicking structured reasoning traces to optimizing executable plans via reinforcement learning.

\noindent \textbf{Stage 1: OOWM Initialization via SFT.}\quad
In the first stage, we utilize the Reasoning-Enhanced Subset (1,000 samples) to initialize the model's ability to ground visual perception into symbolic structures. The primary objective is to teach the model the ``grammar'' of world modeling, treating the expert annotations as a prior distribution for $P(\mathcal{W}|x)$. For each instance, the model is trained to sequentially instantiate the two coupled components of the world model tuple. First, it instantiates the \textbf{State Abstraction ($G_\text{state}$)} within \texttt{<think>} tags, implemented as a UML Class Diagram. This component represents the symbolic Chain-of-Thought (CoT), where the agent explicitly defines object hierarchies and attributes before acting. Subsequently, it generates the \textbf{Control Policy ($G_\text{control}$)} within \texttt{<answer>} tags, implemented as a UML Activity Diagram. This encodes the executable cleaning plan, ensuring that the final output is a logically sound workflow rather than free-form text.

This stage focuses on structural alignment, ensuring the model generates syntactically valid PlantUML code that accurately reflects the expert's object-oriented reasoning process. As demonstrated in Section~\ref{ablation}, a solid structural foundation in SFT is a prerequisite for the success of subsequent GRPO stages.

To rigorously isolate the impact of our structured paradigm, we also prepare baseline variants for comparison: (a) Unstructured CoT \& Plan (Text/Text), and (b) Unstructured CoT with Structured Plan (Text/UML). These variations allow us to quantify the specific contribution of explicitly modeling the environmental state ($\mathcal{S}$) versus strictly modeling the execution policy ($\mathcal{T}$).

\noindent \textbf{Stage 2: Structural Alignment via RLFT.}\quad
In Stage 2, we apply Reinforcement Learning Fine-tuning (RLFT) on the Reasoning-Enhanced Subset (the 1k samples used in SFT). The core objective is to optimize the quality of the \textbf{Control Policy ($G_\text{control}$)}. Crucially, the reward is computed exclusively based on this final executable plan. This design creates a mechanism for latent reward propagation: to maximize the plan's score, the model must implicitly learn to construct a more accurate \textbf{State Abstraction ($G_\text{state}$)} during the reasoning phase. Following SFT, the model has already initialized the capability to instantiate these OOWM components, fulfilling the prerequisites for reinforcement learning.

\begin{figure}[h]
\centering
\includegraphics[width=\linewidth]{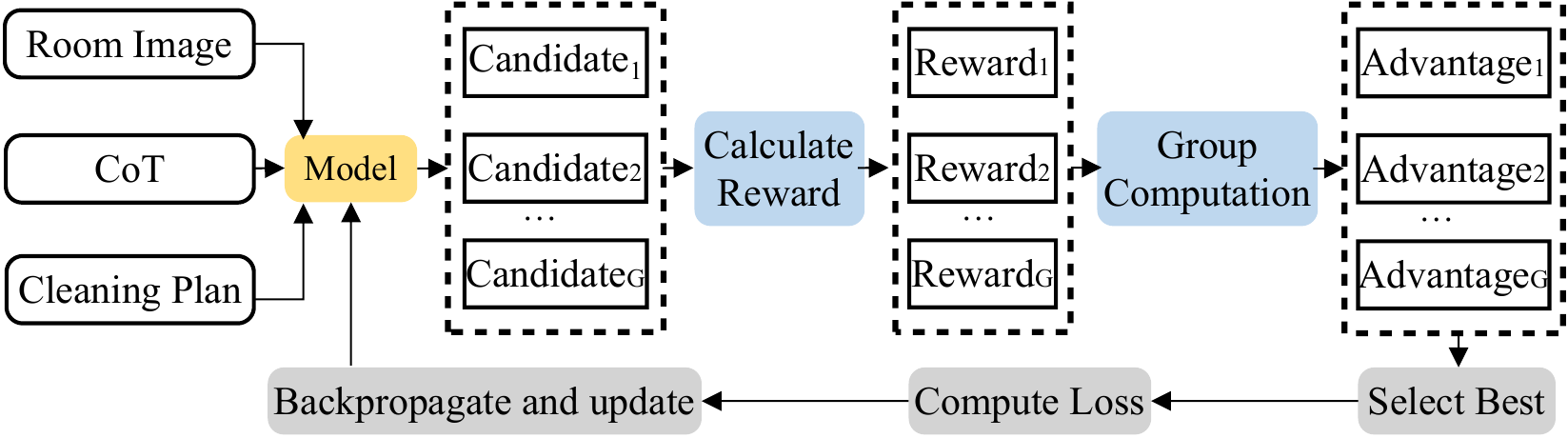} 
\caption{Overview of Group Relative Policy Optimization (GRPO). The framework generates multiple candidate World Model tuples $\mathcal{W} = \langle G_\text{state}, G_{control} \rangle$. It evaluates them using a composite reward that enforces both syntactic correctness and the semantic fidelity of the transition logic, updating parameters to refine the entire reasoning-planning pipeline.}
\label{fig:grpo}
\end{figure}

As shown in Fig. \ref{fig:grpo}, the model receives three inputs: a room image, the ground-truth State Abstraction (CoT), and the reference Control Policy. It generates $G$ plan candidates, and The optimization is driven by a composite reward function: 
\begin{equation}
    r_{total} = r_{struct} + r_{semantic}
\end{equation}

The first component, \textbf{Structural Validity ($r_{struct}$)}, strictly enforces the output format. It assigns a binary score of 1.0 if and only if the generated sequence is correctly encapsulated within the required XML wrappers (\texttt{<think>} and \texttt{<answer>}); otherwise, it assigns 0. The second component, \textbf{Semantic Alignment ($r_{semantic}$)}, measures the logical fidelity of the generated Control Policy ($G_\text{control}$) against the ground truth. Its computation follows a cascaded logic, starting with a Syntax Pre-check. We first verify whether the content within the \texttt{<answer>} tags constitutes a valid PlantUML diagram (i.e., properly enclosed in \texttt{@startuml} ... \texttt{@enduml}). If the diagram syntax is broken or unparseable, $r_{semantic}$ is immediately set to 0. Conversely, if the syntax is valid, we proceed to Content Evaluation. We utilize a stack-based parser to decompose the diagram into three functional partitions: \textbf{Messy Areas}, \textbf{Priority Order}, and \textbf{Specific Steps}. To handle lexical variability, we encode the action nodes in each partition using \texttt{all-MiniLM-L12-v2}. Let $\mathcal{P}_{pred}$ and $\mathcal{P}_{gt}$ be the sets of action nodes in a predicted partition and the ground truth, respectively. We apply a greedy matching algorithm based on cosine similarity to align these sets. The partition-wise reward is calculated as:
\begin{equation}
r_{partition} = \frac{1}{|\mathcal{M}|} \sum_{(i,j) \in \mathcal{M}} \text{sim}(\mathbf{e}_i, \mathbf{e}_j)
\end{equation}
where $\mathcal{M}$ is the set of matched node pairs and $\mathbf{e}$ represents the semantic vector. The final $r_{semantic}$ is the average across all partitions.

Finally, the model parameters are updated using Group Relative Policy Optimization (GRPO). The raw rewards are normalized to compute the advantage:
\begin{equation}
    A_i = \frac{r_i - \mu}{\sigma + \epsilon}
    \label{grpo_advantage}
\end{equation}
where $\mu$ and $\sigma$ are the mean and standard deviation of rewards within the group. The policy loss is defined as:
\begin{equation}
    \mathcal{L}_{GRPO} = -\frac{1}{G} \sum_{i=1}^G \min \left( \frac{\pi_\theta(\hat{y}_i|x)}{\pi_{\theta_{old}}(\hat{y}_i|x)} A_i, \text{clip}(\dots) A_i \right)
    \label{grpo_loss}
\end{equation}

Through this process, the gradient backpropagation from the semantic alignment of $G_{control}$ serves as a verification signal. This effectively refines the upstream $G_\text{state}$, ensuring that the agent's internal perception of the world is optimized to support successful planning.

\begin{algorithm}[htbp]
\footnotesize
\caption{Unified Reward Computation Pipeline for Stage 3}
\label{alg:reward}
\begin{algorithmic}[1]
\Require Predicted output $y_{\text{pred}}$, reference output $y_{\text{ref}}$
\State Initialize $r_{\text{struct}} \gets 0$, $r_{\text{semantic}} \gets 0$

\Comment{\textbf{Step 1: XML Encapsulation Check}}
\If{$y_{\text{pred}}$ contains valid tags \texttt{<think>} and \texttt{<answer>}}
  \State $r_{\text{struct}} \gets 1.0$
\EndIf

\Comment{\textbf{Step 2: Branching based on Target Paradigm}}
\If{Target is \textbf{OOWM (Structured)}}
  \Comment{Requires valid instantiation of Transition Logic ($\mathcal{T}$)}
  \If{$y_{\text{pred}}$ contains valid PlantUML markers \texttt{@startuml} ... \texttt{@enduml}}
    \State Parse $y_{\text{pred}}$ into partitions $P = \{\text{MessyAreas}, \text{Priority}, \text{Steps}\}$
    \ForAll{partition $p \in P$}
      \If{$p$ is successfully extracted}
        \State Encode action nodes in $p$ into vectors $\{\mathbf{e}_i\}$
        \State Compute pairwise similarity matrix with ground truth nodes
        \State Perform greedy matching to find optimal alignment set $\mathcal{M}$
        \State Accumulate similarity: $score_p \gets \frac{1}{|\mathcal{M}|} \sum \text{sim}(i,j)$
      \Else
        \State $score_p \gets 0.0$ \Comment{Missing partition penalty}
      \EndIf
    \EndFor
    \State $r_{\text{semantic}} \gets \text{Mean}(score_{\text{Messy}}, score_{\text{Priority}}, score_{\text{Steps}})$
  \Else
    \State $r_{\text{semantic}} \gets 0.0$ \Comment{Failed OOWM instantiation}
  \EndIf

\ElsIf{Target is \textbf{Baseline (Unstructured)}}
  \Comment{Relaxed document-level comparison}
  \State Encode full text $y_{\text{pred}}$ and $y_{\text{ref}}$ into global vectors $\mathbf{v}_{pred}, \mathbf{v}_{ref}$
  \State $r_{\text{semantic}} \gets \cos(\mathbf{v}_{pred}, \mathbf{v}_{ref})$
\EndIf

\State \Return $\text{Total Reward} = r_{\text{struct}} + r_{\text{semantic}}$
\end{algorithmic}
\end{algorithm}

\noindent\textbf{Stage 3: Scale-Up via Outcome-Based GRPO.} \quad
In the final stage, we scale up the training using the massive \textbf{Base Planning Set} (~29k samples). A critical challenge here is the absence of ground-truth annotations for the State Abstraction ($G_\text{state}$). The dataset provides only the final \textbf{Control Policy ($G_\text{control}$)}. 
Consequently, we treat the State Abstraction ($\mathcal{S}$) as a \textbf{latent variable}. The model must infer a latent $G_\text{state}$ that best supports the generation of the high-reward $G_\text{control}$. By rewarding the semantic fidelity of the final transition logic, the gradient signal propagates backwards through the reasoning chain, implicitly optimizing the agent's internal world modeling process to capture necessary environmental features even without explicit state supervision.

To support both our proposed framework and the unstructured baselines defined in Stage 1, we design a dual-branch reward pipeline:\\
\textbf{For OOWM-based Outputs (Proposed):} We reuse the structural and semantic reward functions from Stage 2. This enforces that the generated plan not only matches the ground truth in content but also strictly adheres to the OOWM serialization syntax (PlantUML) and logical partitioning.\\
\textbf{For Unstructured Baseline Outputs:} To enable fair comparison with text-based approaches, we employ a simplified document-level metric:
\begin{itemize}[leftmargin=*]
    \item \textbf{$r_{struct}$:} 1.0 if both \texttt{<think>} and \texttt{<answer>} tags are present; 0 otherwise.
    \item \textbf{$r_{semantic}$:} We treat the entire cleaning plan as a single unstructured paragraph. We compute the cosine similarity between the predicted text and the reference text using the \texttt{all-MiniLM-L12-\\v2} encoder.
\end{itemize}

This unified mechanism, detailed in Algorithm~\ref{alg:reward}, effectively routes evaluation between the rigorous OOWM structural checks and the flexible baseline text comparisons.

\section{Experiments}
\subsection{Experimental Setup}
\noindent \textbf{Dataset.}\quad
We conduct our experiments on the MRoom-30k benchmark. Consistent with the hierarchical annotation structure defined in Section~\ref{Dataset_construction}, the dataset is utilized as follows: 
i) Reasoning-Enhanced Subset (1k), fully annotated with State Abstraction ($G_\text{state}$), are reserved for initializing the reasoning capability; and 
ii) Base Planning Set (~29k) are randomly split into 80\% for training, 10\% for validation, and 10\% for testing.
Due to computational constraints, we randomly select 2,000 samples from the Base Planning Set for the Stage 3 GRPO fine-tuning. During the final evaluation, we sample a fixed set of 1,000 test instances to assess model performance across all metrics.

\noindent\textbf{Implementation Details.}\quad
The model backbone is based on InternVL 2.5-1B. To rigorously quantify the benefits of our framework, we investigate four distinct input-output configurations, progressing from unstructured text to fully structured world modeling:
\begin{enumerate}[leftmargin=*]
    \item Unstructured Baseline (Text $\to$ Text): A standard VLM-R1 \citep{DBLP:journals/corr/abs-2504-07615} style approach where both the CoT and the cleaning plan are generated as linear natural language. 
    \item Hybrid Strategy (Text $\to$ OOWM): The model uses unstructured textual reasoning and structured Control Policy ($G_{control}$) (serialized as UML). This isolates the benefit of structured output. 
    \item OOWM 2-Stage (OOWM $\to$ OOWM): Our proposed paradigm where both the reasoning trace ($G_\text{state}$) and the plan ($G_{control}$) are structured. This variant is trained without Stage 2. 
    \item OOWM 3-Stage (Full Pipeline): The complete framework, further optimized via RLFT on the Reasoning-Enhanced Subset (Stage 2).
\end{enumerate}
In addition to these internal variants, we compare our method against state-of-the-art prompting strategies, including Tree of Thoughts (ToT) \citep{DBLP:conf/nips/YaoYZS00N23} and Graph of Thoughts (GoT) \citep{DBLP:conf/aaai/BestaBKGPGGLNNH24}.

\noindent\textbf{Evaluation Metrics.}\quad
Conventional n-gram metrics (e.g., ROUGE) are inadequate for assessing the logical validity of cleaning plans. We therefore adopt a Structure-Aware Semantic Evaluation pipeline. The predicted Control Policy and the ground truth are first decomposed into their functional partitions. We then perform node-level alignment using a similarity matrix computed via all-MiniLM-L12-v2 embeddings. We report two types of metrics:
i) Semantic Fidelity (Regression): The average cosine similarity across all matched node pairs, measuring how closely the generated actions resemble the expert's intent; and 
ii) Execution Statistics (Classification): Using a fixed similarity threshold (0.5), we classify nodes as True Positives (TP), False Negatives (FN), or False Positives (FP). We compute Precision, Recall, and F1-score. Notably, we interpret Recall as the Task Execution Success Rate, as it measures the proportion of necessary ground-truth actions successfully recovered by the agent's policy. To ensure consistent evaluation, textual instructions are converted into UML activity diagrams using GPT-4o before scoring.

\begin{table*}[h]
\centering
\caption{Quantitative comparison on the MRoom-30k benchmark. We report semantic similarity and execution metrics. Note that \textbf{Recall} serves as a proxy for the \textbf{Task Execution Success Rate}. The best results are highlighted in \textbf{bold}, and the second-best are \underline{underlined}.}
\label{tab:cot_plan_results}
\begin{tabular}{llcccc} % 移除了多余的一列
\toprule
 & \textbf{Method} & \textbf{Similarity} & \textbf{Precision} & \textbf{Recall (Success Rate)} & \textbf{F1} \\
\midrule
\multirow{3}{*}{Baselines} 
  & Tree of Thoughts \citep{DBLP:conf/nips/YaoYZS00N23}   & 0.4209 & 0.4854          & 0.4639 & 0.4695 \\
  & Graph of Thoughts \citep{DBLP:conf/aaai/BestaBKGPGGLNNH24}  & 0.5383 & 0.5263          & 0.5579 & 0.5371 \\
  & Unstructured Baseline (Text $\to$ Text) \citep{DBLP:journals/corr/abs-2504-07615}     & 0.5498 & \textbf{0.5489} & 0.6280 & 0.5811 \\
\midrule
\multirow{3}{*}{\textbf{OOWM (Ours)}}
  & Hybrid Strategy (Text $\to$ OOWM)      & 0.5562 & \underline{0.5384} & 0.6438 & \underline{0.5812} \\
  & OOWM 2-Stage (OOWM $\to$ OOWM)       & \underline{0.5617} & 0.5304 & \underline{0.6536} & 0.5803 \\
  & OOWM 3-Stage (Full Pipeline)       & \textbf{0.5694} & 0.5326 & \textbf{0.6744} & \textbf{0.5904} \\
\bottomrule
\end{tabular}
\end{table*}

\subsection{Evaluation Results}
Beyond training dynamics, we benchmark our method against state-of-the-art unstructured approaches—including standard Text-based CoT \citep{DBLP:journals/corr/abs-2504-07615}, Tree of Thoughts \citep{DBLP:conf/nips/YaoYZS00N23}, and Graph of Thoughts \citep{DBLP:conf/aaai/BestaBKGPGGLNNH24}—with quantitative results summarized in Table~\ref{tab:cot_plan_results}.

\noindent\textbf{The State Representation Gap in Baselines.}\quad
Notably, while the Unstructured Baseline (Text-CoT) achieves the highest precision (0.5489), it lags significantly in recall and overall F1. This performance skew reveals a fundamental flaw in text-only reasoning: without an explicit State Abstraction ($\mathcal{S}$), the model lacks a persistent memory buffer to track object states. Consequently, it adopts a conservative strategy—generating fewer, safe steps but failing to capture the comprehensive set of actions required for task completion. Similarly, complex prompting strategies like Tree of Thoughts and Graph of Thoughts demonstrate lower performance across key metrics. This indicates that increasing the complexity of textual reasoning without introducing object-oriented formalism is insufficient for robust embodied planning; the model simply hallucinates more elaborate but structurally unsound plans.

\noindent\textbf{Benefit of Transition Logic Constraints.}\quad
By contrast, the Hybrid Strategy (Text $\to$ OOWM), which forces the output into a structured Control Policy ($G_\text{control}$), results in immediate improvements in recall and similarity. This suggests that the rigorous syntax of the Activity Diagram acts as a behavioral scaffold. Even when the upstream reasoning remains unstructured, the requirement to instantiate a valid Transition Logic ($\mathcal{T}$) compels the model to generate more systematic and logically complete workflows, reducing the omission of critical steps.

\noindent\textbf{Superiority of the Full OOWM Paradigm.}\quad
The most significant gains are realized when the full World Model pipeline is instantiated. The OOWM 2-Stage model, which grounds its planning in an explicit State Abstraction ($G_\text{state}$), consistently outperforms the Hybrid Strategy. This validates our core hypothesis: symbolic object-oriented reasoning ($\mathcal{S}$) is inherently more compatible with executable planning than free-form text.
Ultimately, the OOWM 3-Stage configuration achieves peak performance, recording the highest semantic similarity (0.5694), recall (0.6744), and F1 score (0.5904). This confirms the efficacy of our latent reward propagation mechanism: by optimizing the downstream execution policy via GRPO, the model implicitly refines its internal state abstraction, yielding a cleaning policy that is not only structurally valid but semantically grounded in the physical environment.

\subsection{Training Dynamics}
We analyze the training stability and reward convergence of three distinct modeling paradigms during the Stage 3 GRPO phase:
i) \textbf{Hybrid Strategy (w/o Stage 2)}, which bypasses explicit State Abstraction ($G_\text{state}$);
ii) \textbf{OOWM Direct (w/o Stage 2)}, which instantiates the full tuple but lacks structural pre-alignment; and
iii) \textbf{OOWM Full-Stage (w/ Stage 2)}, which incorporates the full latent reward propagation pipeline.

\begin{figure}[t]
  \centering
  \subfloat[Structural Validity]{%
    \includegraphics[width=0.48\linewidth]{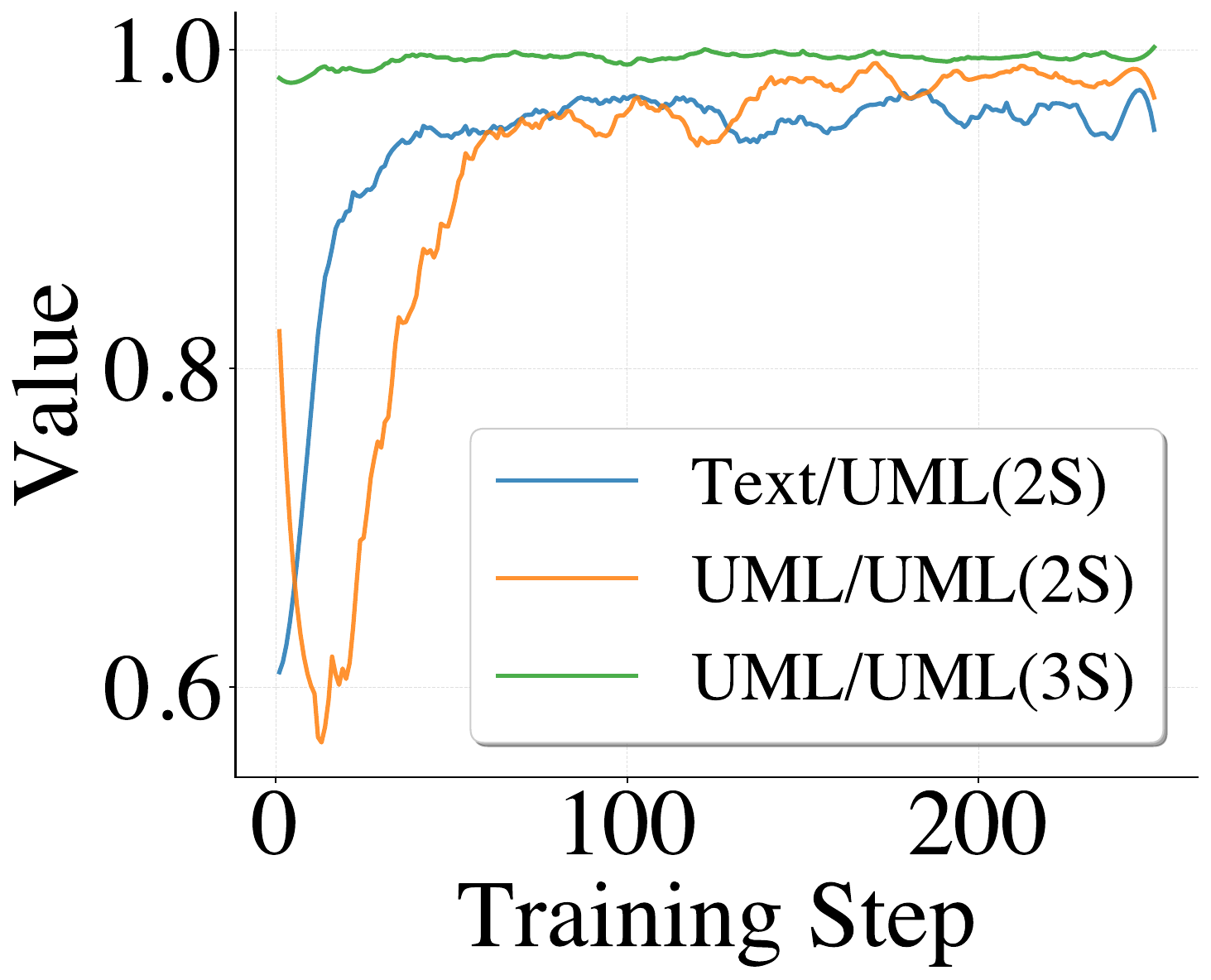}\label{fig:format_reward}}
  \hfill
  \subfloat[Semantic Alignment]{%
    \includegraphics[width=0.48\linewidth]{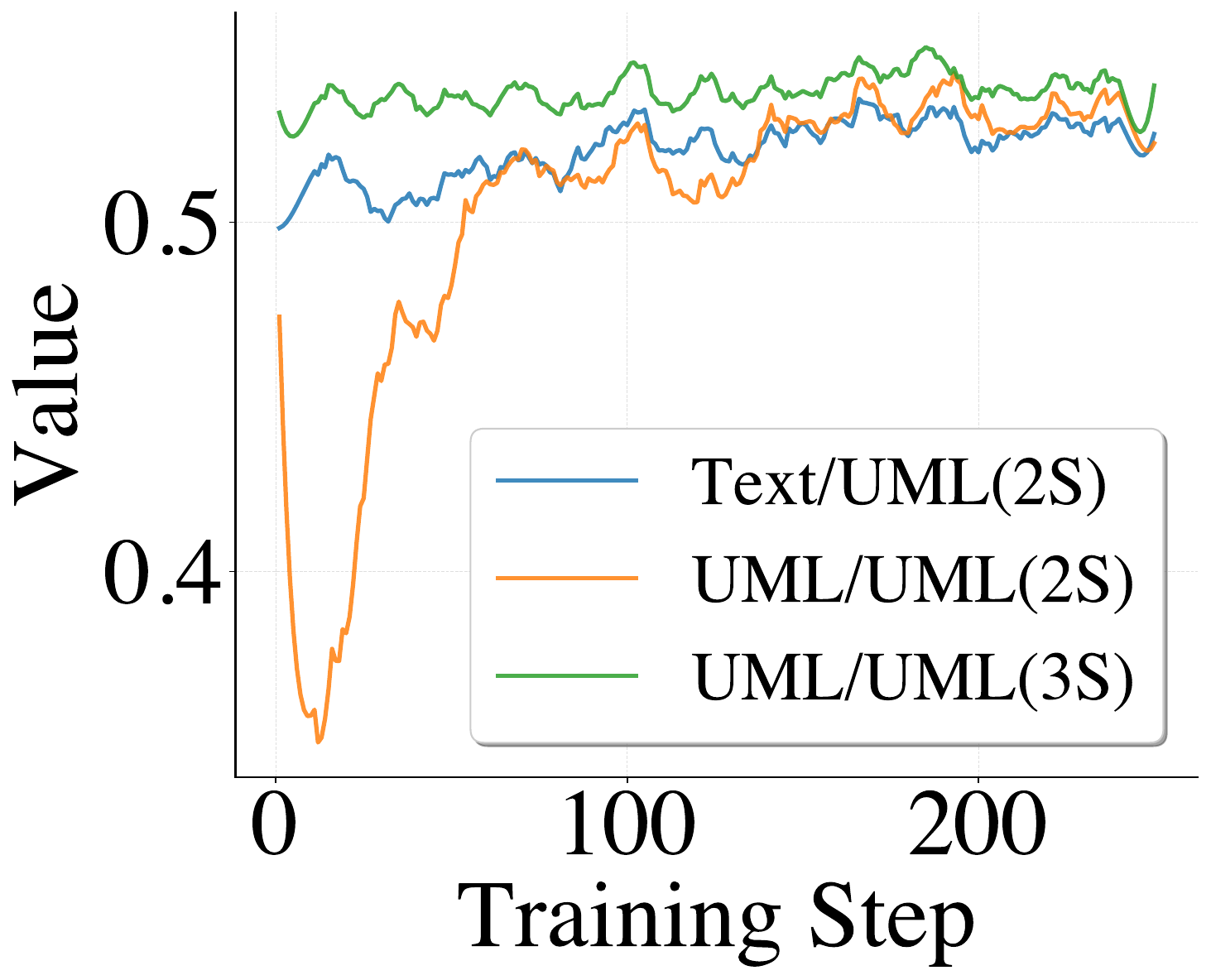}\label{fig:accu_reward}}
   
  \subfloat[Total Reward]{%
    \includegraphics[width=0.48\linewidth]{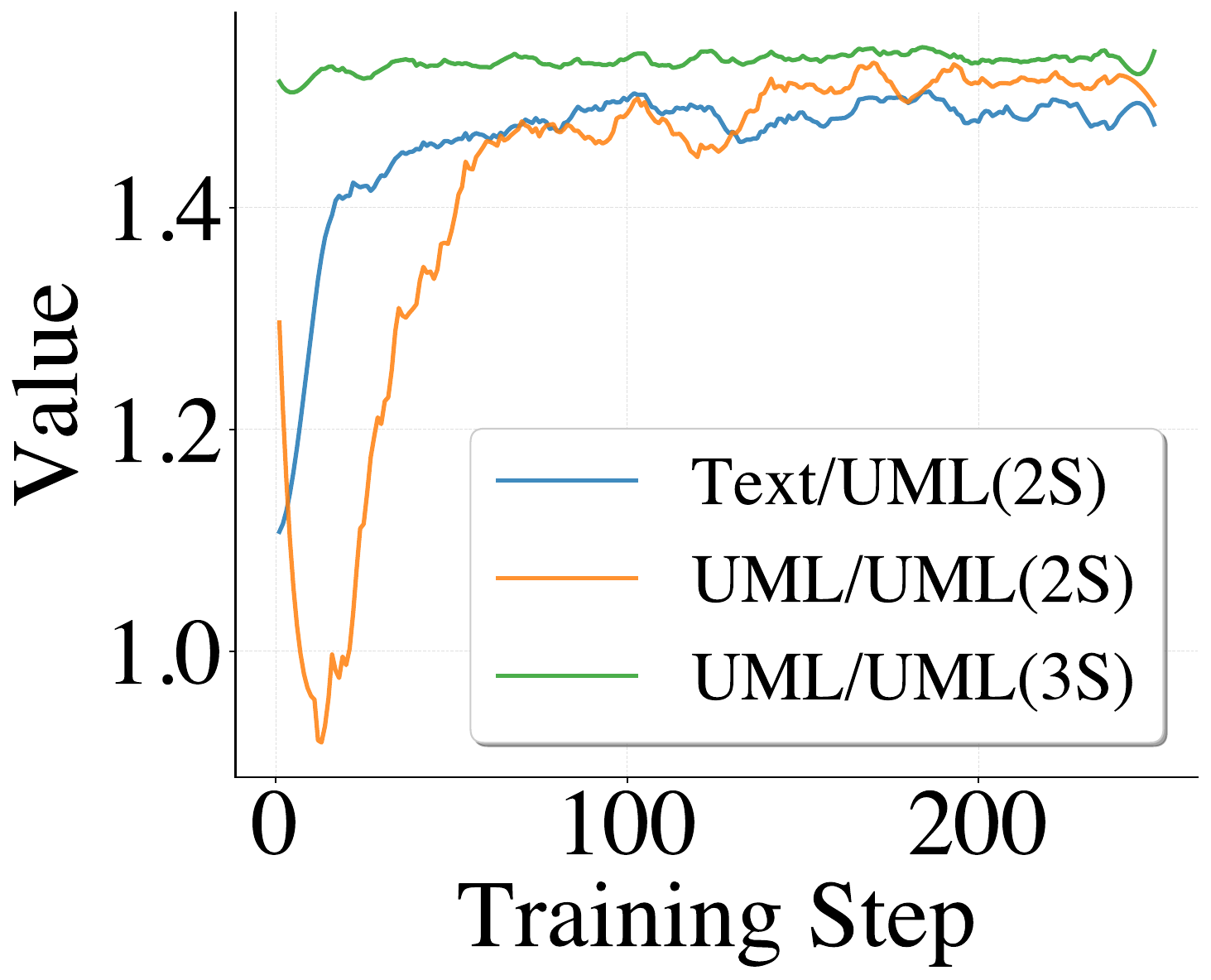}\label{fig:total_reward}}
  \hfill
  \subfloat[Training Loss]{%
    \includegraphics[width=0.48\linewidth]{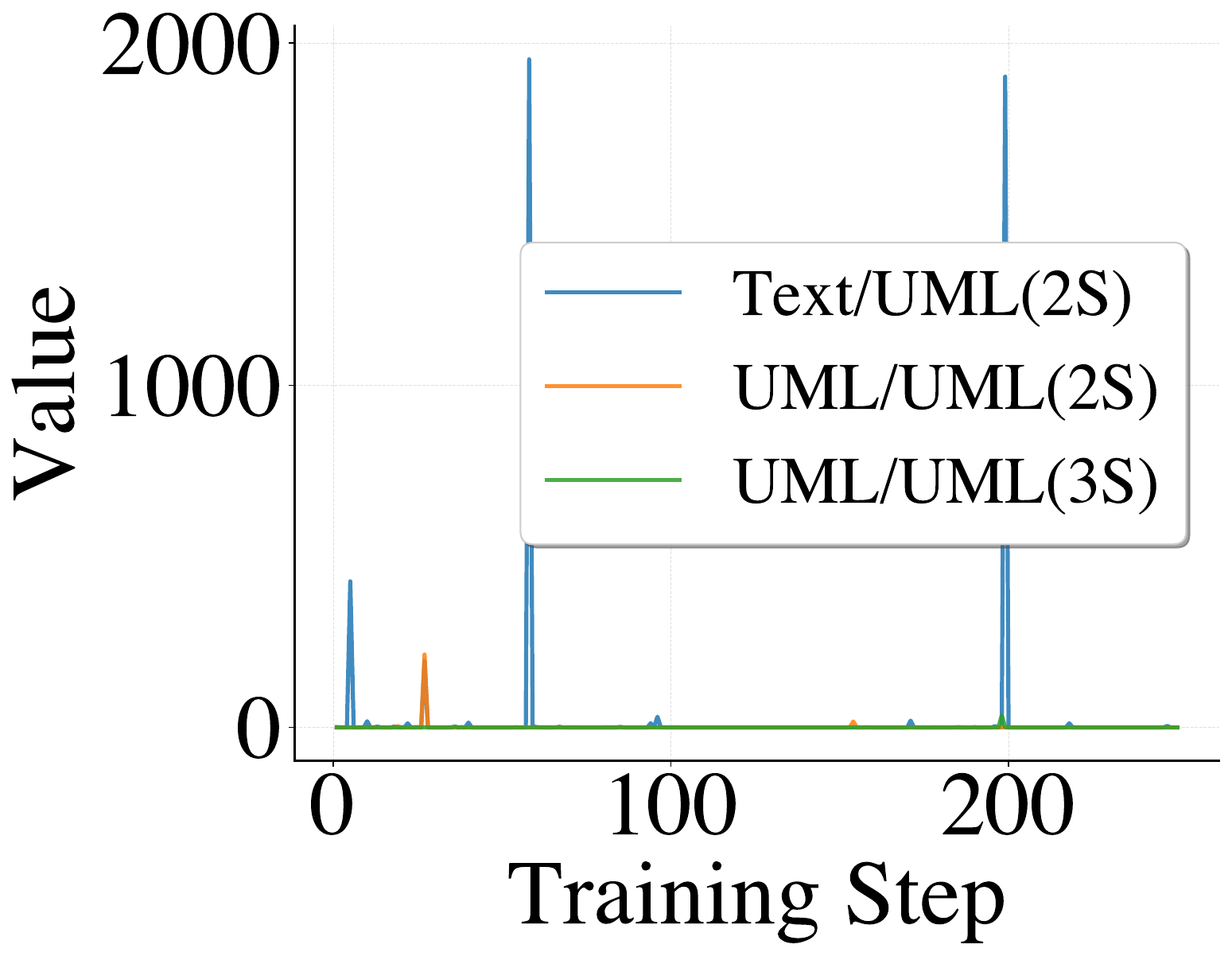}\label{fig:loss_curve}}
    
  \caption{Comparison of training dynamics during Stage 3 GRPO. 
  (a) \textbf{Structural Validity ($r_{struct}$)} measures OOWM instantiation success. 
  (b) \textbf{Semantic Alignment ($r_{semantic}$)} evaluates the logical fidelity of the Transition Logic ($G_{control}$). 
  (c) Total Reward evolution. 
  (d) Training Loss convergence. 
  The \textbf{OOWM Full-Stage} model (green) demonstrates superior stability and asymptotic performance.}
  \label{fig:overall_metrics}
\end{figure}

\noindent\textbf{Cost and Benefit of Explicit Modeling.}\quad
As shown in Fig.~\ref{fig:overall_metrics} (a-c), \textbf{OOWM Direct} exhibits a ``slow-start, high-ceiling'' trajectory. The initial lag stems from the modeling burden: the agent must learn to ground visual inputs into a structured State Abstraction ($\mathcal{S}$) before effectively optimizing the policy. However, the subsequent crossover surpasses the Hybrid Strategy, confirming that $G_\text{state}$ acts as a cognitive regularizer. Unlike the loose text-to-policy associations in Hybrid models, the object-oriented formalism effectively prunes the search space for the Transition Logic ($\mathcal{T}$), enabling the discovery of more logically consistent plans in the long run.

\noindent\textbf{Impact of Latent Reward Propagation.}\quad
The \textbf{OOWM Full-Stage} model demonstrates the most significant gains, characterized by rapid convergence and minimal loss variance (Fig.~\ref{fig:loss_curve}). This validates Stage 2 as a critical Structural Alignment phase. By pre-optimizing the consistency between $\mathcal{S}$ and $\mathcal{T}$, the model enters Stage 3 with a ``warm-starte'' world model. Consequently, outcome-based gradient signals refine an already grounded structure rather than constructing one from scratch, preventing the optimization instability observed in unaligned baselines.

\subsection{Ablation Study}
\label{ablation}

\begin{figure}[h]
  \centering
  \begin{minipage}[t]{0.48\linewidth}
    \centering
    \includegraphics[width=\linewidth]{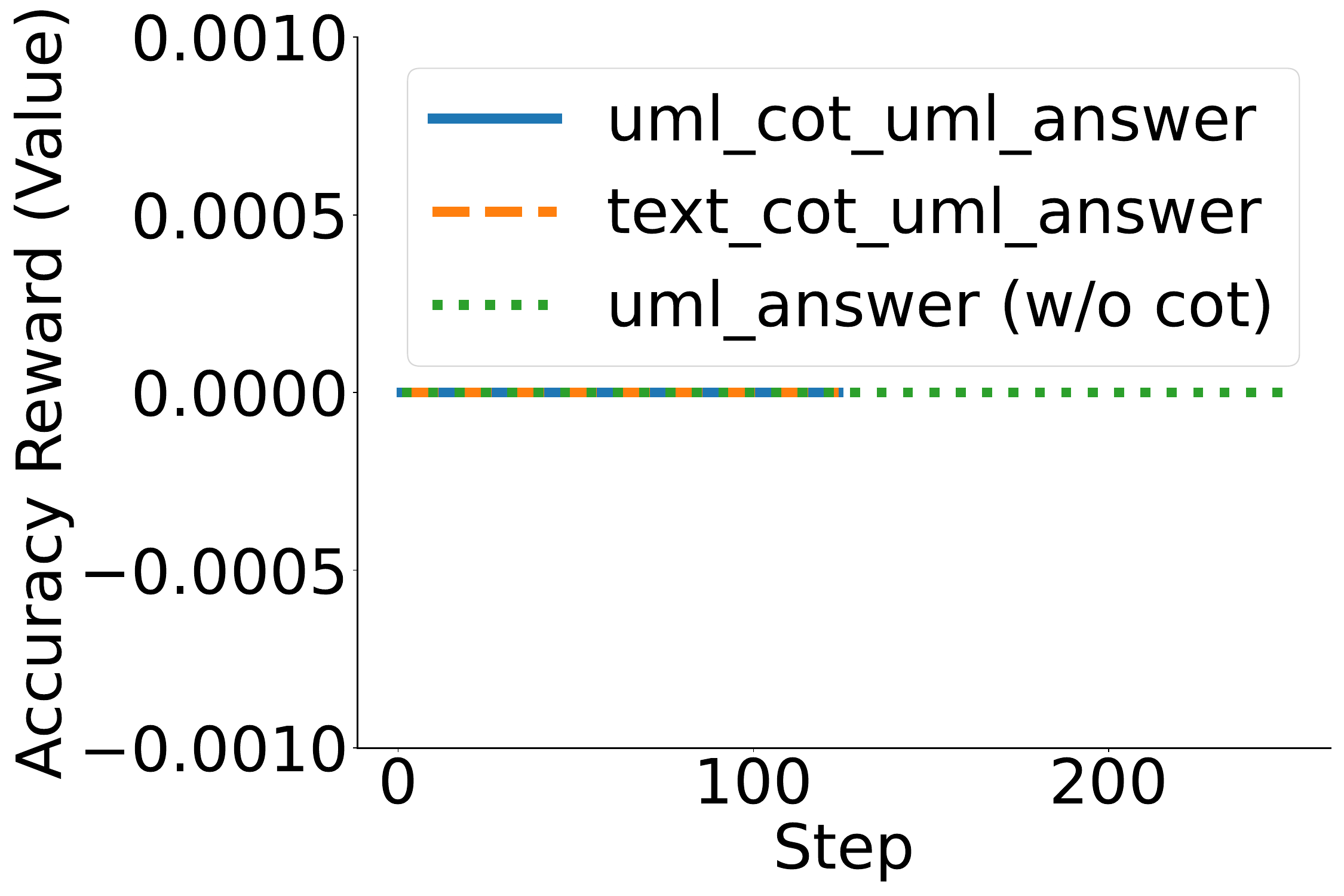}
    \captionof{figure}{Impact of SFT initialization on OOWM instantiation success.}
    \label{fig:sft_initialization}
  \end{minipage}
  \hfill
  \begin{minipage}[t]{0.48\linewidth}
    \centering
    \includegraphics[width=\linewidth]{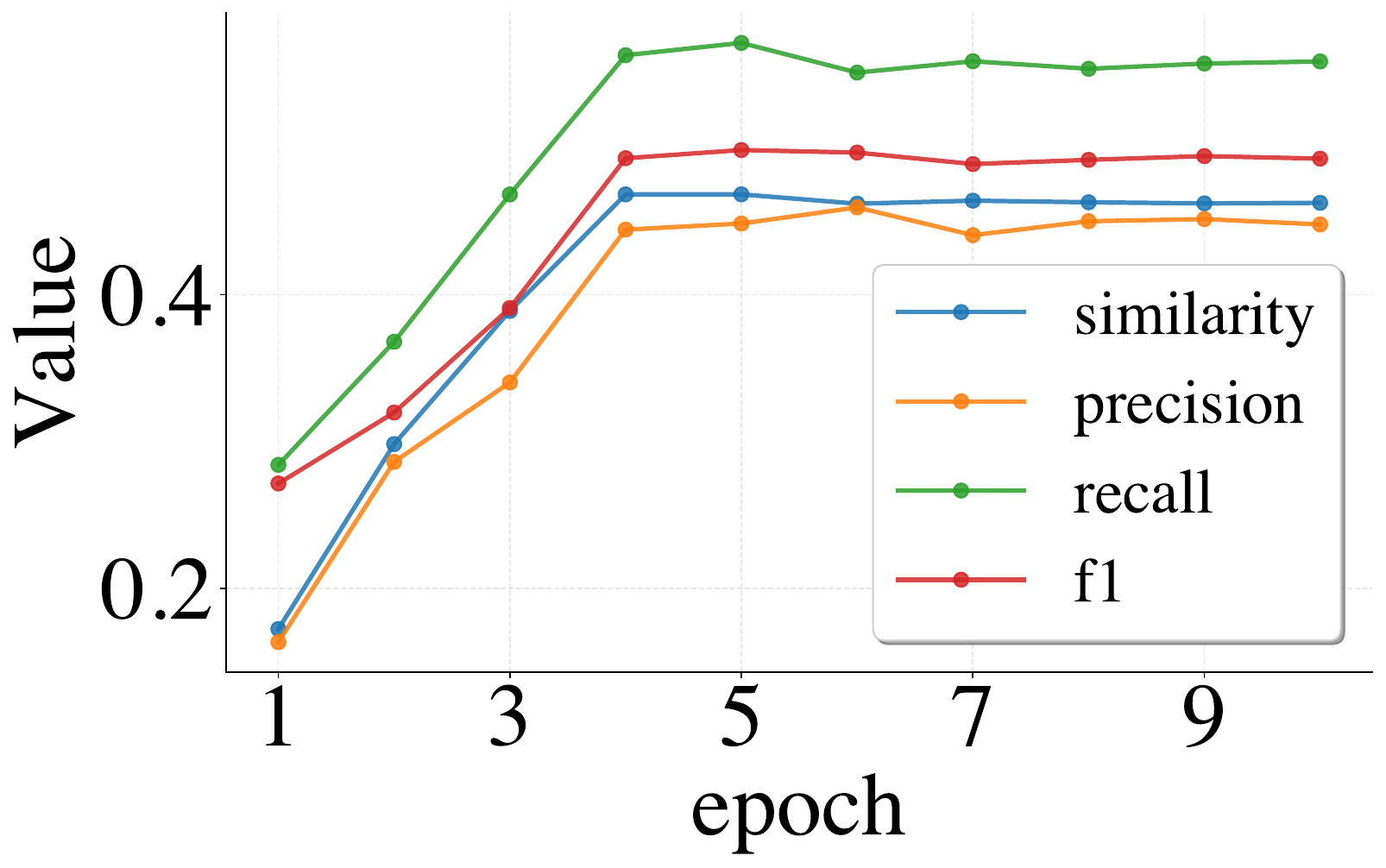}
    \captionof{figure}{Performance saturation of SFT vs. GRPO optimization.}
    \label{fig:sft_metric_curve}
  \end{minipage}
\end{figure}

\noindent\textbf{Necessity of Structural Bootstrapping (SFT).}\quad
We first investigate whether the agent can learn to instantiate the World Model tuple $\mathcal{W} = \langle \mathcal{S}, \mathcal{T} \rangle$ solely through reinforcement learning. We initialize the model without Stage 1 supervision and apply Stage 3 GRPO directly on the Base Planning Set.
The results (Fig.~\ref{fig:sft_initialization}) show a complete failure to converge. This confirms that the search space for valid PlantUML syntax is too sparse for random exploration. Without the prior distribution provided by SFT, the model cannot generate the correct PluntUML grammar required to trigger the Semantic Alignment reward ($r_{semantic}$). Thus, SFT serves as a critical structural bootstrapping phase: it teaches the agent the ``grammar'' of the OOWM formalism, a strict prerequisite for any subsequent semantic refinement.

\noindent\textbf{Beyond Mimicry: GRPO vs. Extended SFT.}\quad
To determine whether the gains in Stage 3 stem from the specific optimization algorithm or merely extended training, we compare the performance trajectories of continued SFT versus switching to GRPO.
As shown in Fig.~\ref{fig:sft_metric_curve}, pure SFT metrics (Precision, Recall, F1) plateau rapidly after epoch 5. This indicates that supervised imitation has reached a saturation point: the model masters the syntactic form of the components but struggles to further refine the underlying Transition Logic ($\mathcal{T}$) solely by minimizing token prediction error.
In contrast, applying GRPO from the epoch 5 checkpoint breaks this ``imitation ceiling,'' yielding consistent improvements across all metrics. This validates the efficacy of our outcome-based optimization. Unlike SFT, which mimics the expert's static traces, GRPO rewards the semantic utility of the final plan. This forces the model to implicitly adjust its latent State Abstraction ($\mathcal{S}$) to maximize the correctness of the generated policies, resulting in a more robust and actionable world model.

\subsection{Cross-Task Generalization}
We evaluate the adaptability of our Object-Oriented World Modeling (OOWM) paradigm by testing it on two unseen domains: \textit{Cooking} and \textit{Painting}. These tasks require the agent to instantiate new object hierarchies and action sequences without prior domain-specific training.

\begin{table}[h]
\centering
\footnotesize
\caption{Quantitative evaluation of cross-task generalization on Cooking and Painting domains. We report Similarity, Precision, Recall, and F1 scores for different model configurations.}
\label{tab:generalization}
\begin{tabular}{c l r r r r}
\toprule
Task & Model & Similarity & Precision & Recall & F1 \\
\midrule
\multirow{4}{*}{Cooking}
 & Unstructured Baseline & 0.6357 & 0.2010 & 0.4219 & 0.2694 \\
 & Hybrid Strategy  & 0.5705 & 0.2683 & 0.4269 & 0.3213 \\
 & OOWM 2-Stage   & 0.6058 & 0.4119 & 0.4059 & 0.3076 \\
 & OOWM 3-Stage   & 0.6889 & 0.3447 & 0.4448 & 0.3824 \\
\midrule
\multirow{4}{*}{Painting}
 & Unstructured Baseline & 0.6040 & 0.1471 & 0.3665 & 0.2087 \\
 & Hybrid Strategy  & 0.5750 & 0.1750 & 0.1555 & 0.1643 \\
 & OOWM 2-Stage   & 0.6156 & 0.1566 & 0.1498 & 0.1531 \\
 & OOWM 3-Stage   & 0.6503 & 0.1892 & 0.2769 & 0.1715 \\
\bottomrule
\end{tabular}
\end{table}

In the \textbf{Cooking} task, OOWM 3-Stage achieves superior Similarity and F1, validating the transferability of the $G_\text{state} \to G_\text{control}$ meta-structure. Notably, OOWM 2-Stage attains the highest Precision, proving that explicit object typing effectively reduces hallucinations—a key weakness in the Unstructured Baseline.

The \textbf{Painting} task proves significantly harder. While OOWM 3-Stage retains the lead in Similarity and Precision, the Unstructured Baseline yields higher Recall, likely due to generic descriptions maximizing lexical overlap. In contrast, OOWM attempts to construct rigorous models but fails when facing unseen objects. This indicates that while OOWM offers a strong reasoning scaffold, its cross-domain success remains bounded by the backbone's visual grounding capabilities.

\section{Conclusion}
In this work, we introduce Object-Oriented World Modeling (OOWM), a framework that fundamentally redefines embodied reasoning not as linear text generation, but as symbolic system design. By formally coupling a State Abstraction ($\mathcal{S}$, instantiated as $G_\text{state}$) with a Transition Logic ($\mathcal{T}$, instantiated as $G_\text{control}$), we bridge the gap between high-level semantic understanding and low-level executable control. 
Our progressive three-stage training strategy (SFT, RLFT, and GRPO) proves critical to this framework. Specifically, the outcome-based reinforcement learning efficiently optimizes the underlying world model through latent reward propagation, ensuring that the agent's internal representation supports robust decision-making. 
Extensive experiments on the MRoom-30k benchmark demonstrate that the OOWM paradigm significantly outperforms unstructured textual baselines and hybrid approaches. These results confirm that imposing structural architectural constraints is a more effective path toward robust embodied agents than relying on unstructured reasoning alone.

\bibliographystyle{ACM-Reference-Format}
\bibliography{sample-base}

\end{document}